\documentclass[]{ceurart}
\begin{document}

\copyrightyear{2024}
\copyrightclause{Copyright for this paper by its authors.
  Use permitted under Creative Commons License Attribution 4.0
  International (CC BY 4.0).}

\conference{CLEF 2024: Conference and Labs of the Evaluation Forum, September 9-12, 2024, Grenoble, France}

\title{Transfer Learning with Self-Supervised Vision Transformers for Snake Identification}

\author[1]{Anthony Miyaguchi}[
orcid=0000-0002-9165-8718,
email=acmiyaguchi@gatech.edu,
]
\cormark[1]
\author[1]{Murilo Gustineli}[
email=murilogustineli@gatech.edu,
]
\cormark[1]
\author[1]{Austin Fischer}[
email=afischer39@gatech.edu,
]
\author[1]{Ryan Lundqvist}[
email=rlundqvist3@gatech.edu,
]

\address[1]{Georgia Institute of Technology, North Ave NW, Atlanta, GA 30332}
\cortext[1]{Corresponding author.}

\begin{abstract}
    We present our approach for the SnakeCLEF 2024 competition to predict snake species from images.
    We explore and use Meta's DINOv2 vision transformer model for feature extraction to tackle species' high variability and visual similarity in a dataset of 182,261 images. 
    We perform exploratory analysis on embeddings to understand their structure, and train a linear classifier on the embeddings to predict species.
    Despite achieving a score of 39.69, our results show promise for DINOv2 embeddings in snake identification. 
    All code for this project is available at \url{https://github.com/dsgt-kaggle-clef/snakeclef-2024}.
\end{abstract}


\maketitle

\section{Introduction}

SnakeCLEF \cite{snakeclef2024} is a task in the LifeCLEF lab \cite{lifeclef2024} that prompts the development of a classification system for identifying venomous and harmless species of snakes from images.
The dataset includes 103,404 observations with 182,261 images of 1,784 species from 214 countries, sourced from the iNaturalist and Herpmapper platforms. 
19.5\% of the species are venomous, and the dataset is imbalanced.
The dataset poses significant challenges in classification due to the high variability in species found in any particular geographical location and the relatively high degree of visual similarity between species due to mimicry and convergence.

\section{Self-Supervised Vision Transformers}

Vision transformers (ViTs) treat images as a regular grid of smaller patches processed sequentially, like words in a sentence.
Semantic information is embedded into a high-dimensional space that preserves semantic information.
The attention mechanism of transformers can learn to optimize patterns across a sequence of embedded patch tokens. 
The basic method for self-supervision on images involves randomly masking patches in an auto-encoder architecture, which enables learning in large data regimes effectively \cite{he2022masked}.
Vision transformers have demonstrated state-of-the-art performance, overtaking convolutional neural networks (CNN) in the computer vision domain.

DINOv2 \cite{oquab2023dinov2} is a self-supervised vision transformer developed by Meta.
DINOv2 demonstrates strong out-of-distribution performance, making it ideal for dealing with various previously unseen snake photographs in varied scenarios.
The DINOv2 architecture is available in various parameterizations with distinctive patch token dimensions: small (S) with 382, base (B) with 768, large (L) with 1024, and giant (g) with 1536.
The model is trained using LVD-142M, a massive collection of images that combines other datasets such as ImageNet-22k.
The base model generates embeddings with a fixed size of $\mathbb{R}^{257 \times 768}$, regardless of the input image dimensions.
This is achieved by dividing each image into 256 fixed-size patches with an additional [CLS] token in $\mathbb{R}^{768}$.

\section{Exploratory Data Analysis}

\begin{figure}[t]
    \centering
    \includegraphics[width=0.49\textwidth]{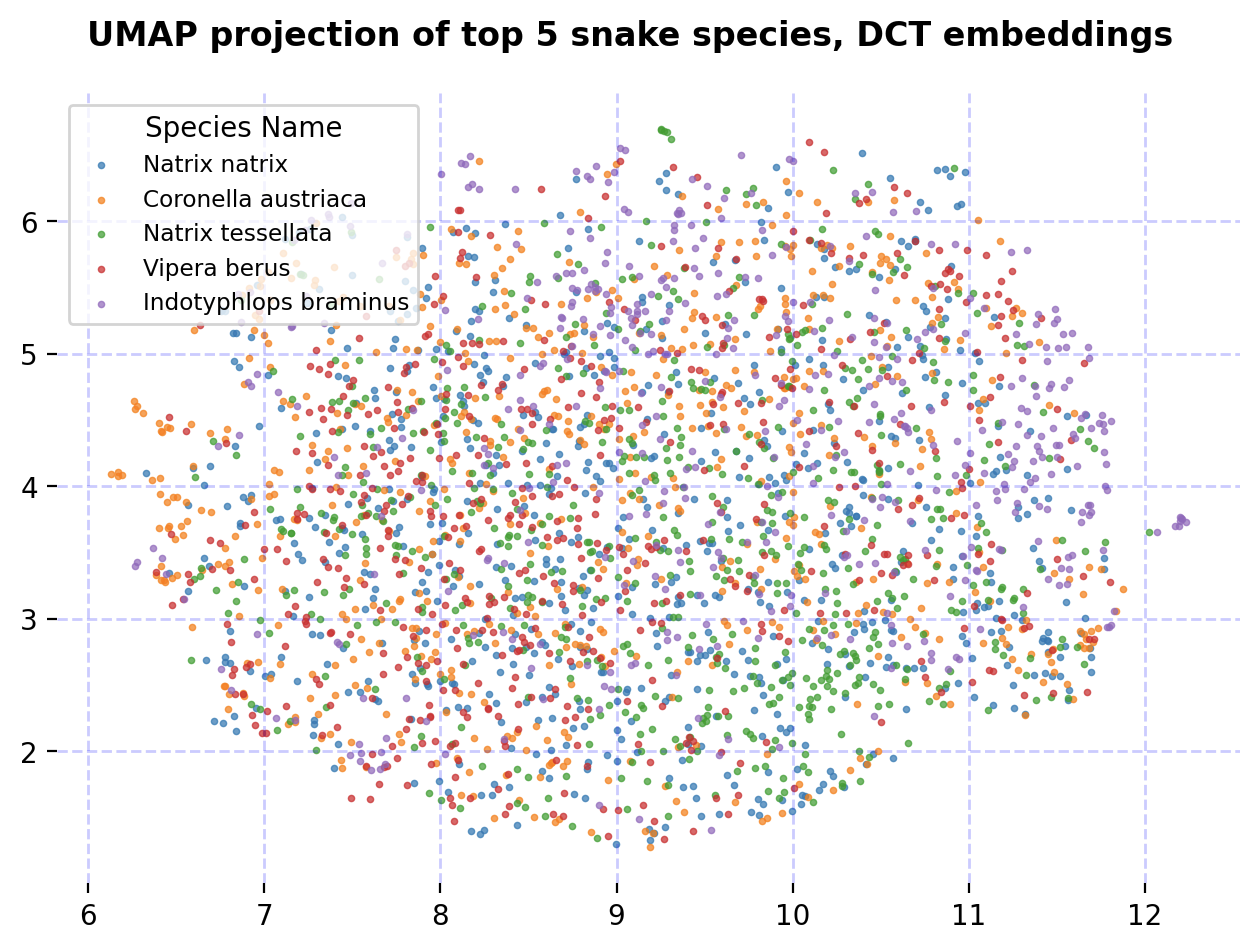}
    \includegraphics[width=0.49\textwidth]{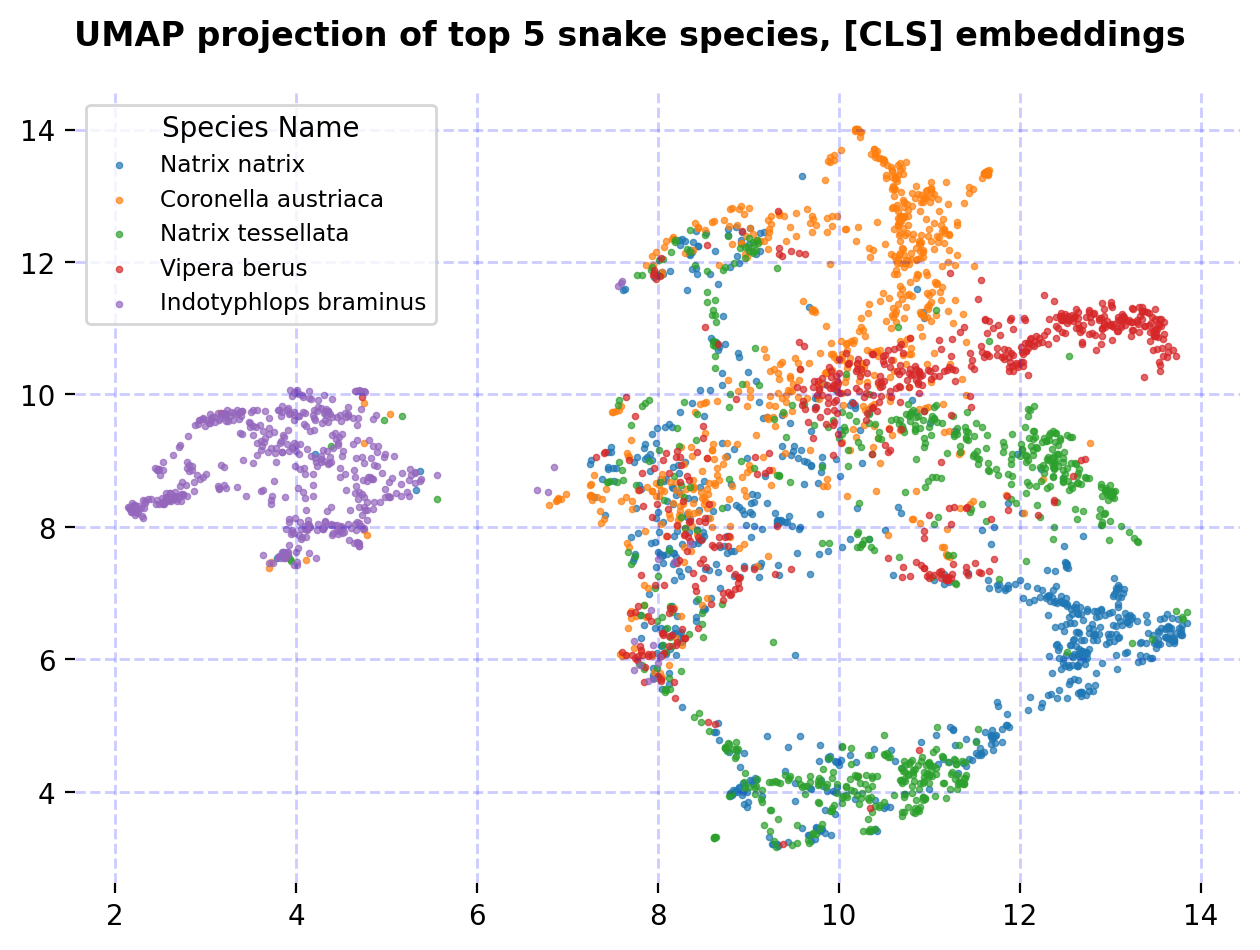}
    \caption{
    UMAP projection of DCT and [CLS] embeddings of the top 5 snake species by image count.
    }
  \label{fig:top5-species}
\end{figure}

The DINOv2 embeddings are learned representations of the data that map images into a manifold, preserving geometrical notions of distance from the input space. 
The quality and differentiability of these outputs are critical for downstream modeling.
We perform dimensionality reduction of the embeddings to compress the data in two ways: by extracting the top $8 \times 8$ coefficients of the 2D discrete cosine transform (DCT) on the $\mathbb{R}^{256 \times 768}$ patch tokens, and the [CLS] token in $\mathbb{R}^{768}$.
The DCT effectively captures periodic patterns in perceptual domains such as images and audio, so we explore its behavior as a computationally efficient data-independent lossy compression technique.
The [CLS] token is a discriminator in the self-supervision process of training the ViT, and thus, it should also serve as a good representation of how distant points are in space.
Transfer learning tasks often use the [CLS] token from vision transformers to train classifiers on domain-specific tasks \cite{oquab2023dinov2}. 

\begin{figure}[t]
    \centering
    \includegraphics[width=0.75\textwidth]{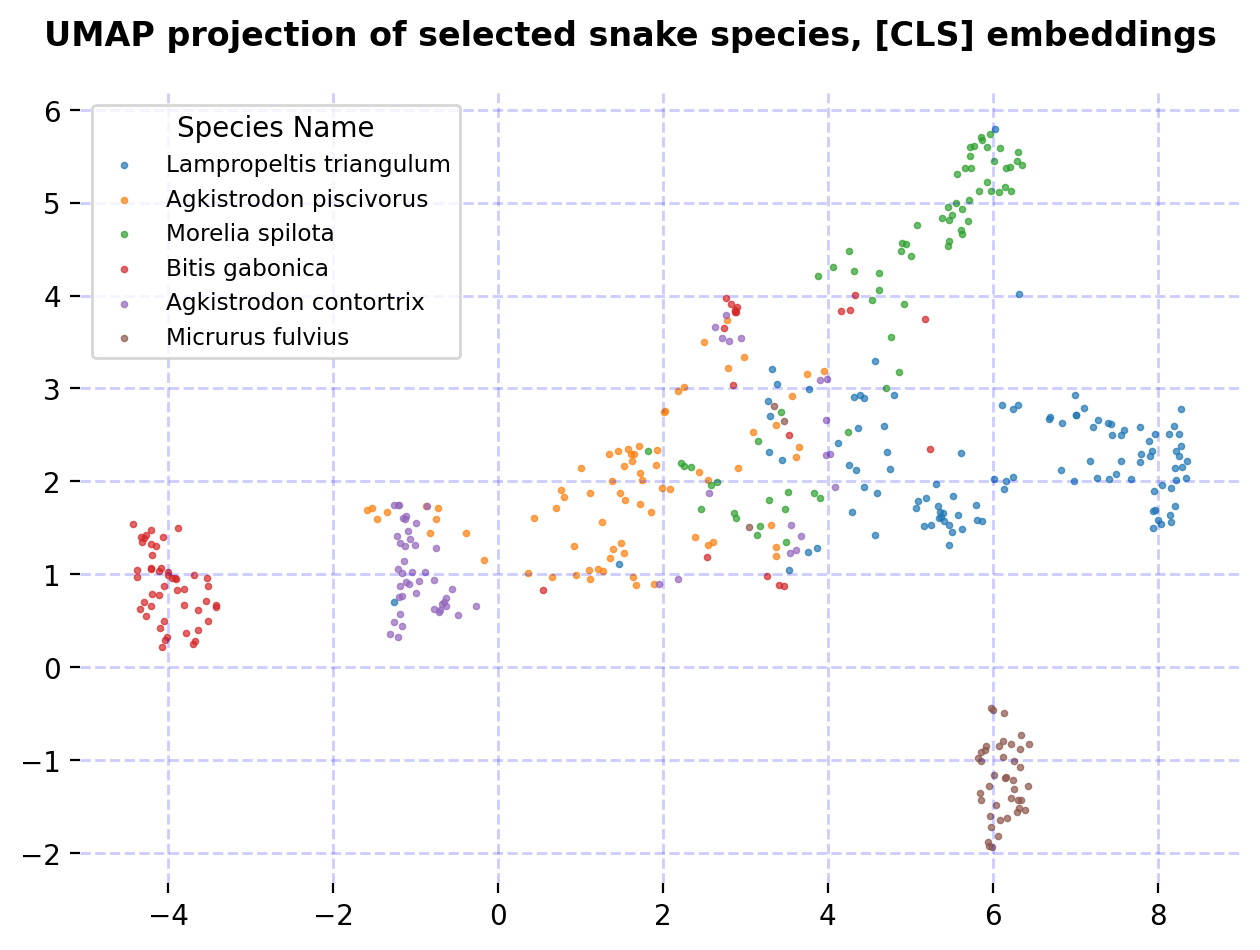}
    \caption{
    A selected subset of snake species with unique features relevant to the task.
    We extracted the [CLS] token embeddings using DINOv2 and created projections via UMAP.
    The distribution is expected, with species like Agkistrodon piscivorus and Agkistrodon contortrix being represented similarly. Additionally, species that are more biologically polymorphic and differ in appearance regionally, like Lampropeltis triangulum and Morelia spilota, exhibit a wider spread. 
    In contrast, species that appear more uniform, like Bitis gabonica and Micrurus fulvius, are found in distinct clusters. 
    }
  \label{fig:selected-species}
\end{figure}

We visually inspect embeddings by reducing their dimension and scatter plotting the results in two dimensions using class labels as colors.
We expect similar images to cluster since the projected manifolds learn a conceptual distance between points.
In Figure \ref{fig:top5-species} we plot UMAP \cite{mcinnes2018umap} projections of these embeddings, both their [CLS] tokens and the DCT coefficients of the patch tokens.
[CLS] tokens demonstrate better clustering between species than the DCT output, which has little apparent coherence in their 2D projections.
The 2D DCT is either filtered too harshly (i.e., 64 coefficients are not enough for fine details in the landscape) or there is little periodicity in the spectral structure of the tokens to take advantage of.
The [CLS] tokens can be seen to cluster distinctly for the images of some species, such as Indotyphlops Braminus in Figure \ref{fig:top5-species} and Bitis Gabonica in Figure \ref{fig:selected-species}. 
There appear to be nodes for most species, but some pairs of species have highly interspersed distributions, such as Coronella Austriaca and Vipera Berus in Figure \ref{fig:top5-species}. 
The rich structure expressed in the 2D projection of the embedding signals a representation amenable to tasks in other domains.

Given the ability to visually discriminate between species images through a clustering analysis, we construct a dataset containing a subset of six species that may represent challenges posed by the competition in Figure \ref{fig:selected-species}.
The first selected species, Micrurus fulvius, is a coral snake belonging to Elapidae and thus has medically significant venom. 
The second selected snake, Lampropeltis triangulum, or the non-venomous milk snake, is a member of Colubridae and is known to be visually similar to venomous coral snakes due to Batesian mimicry. 
Effectively classifying these two snakes is significant for the mission of the competition and for improving human health. 
The third selected species is Morelia spilota, the carpet python, which diversifies the subset of data to include individuals belonging to Pythonidae and has ample associated individual images in the dataset.
Finally, we examine 3 venomous thick-bodied vipers with similar life strategies and patterning as leaf litter ambush predators, being: gaboon vipers (Bitis gabonica), copperheads (Agkistrodon contortrix), and cottonmouths (Agkistrodon piscivorus). 
These are beneficial to study as they represent Viperidae and vary in geographical range, with Gaboon vipers residing in sub-Saharan Africa and the latter residing in the contiguous United States. 
Additionally, differentiating between young copperheads and young cottonmouths can prove challenging yet essential, with cottonmouths possessing more medically significant venom. 

\section{Methodology}

\begin{figure}[ht]
    \centering
    \includegraphics[width=\textwidth]{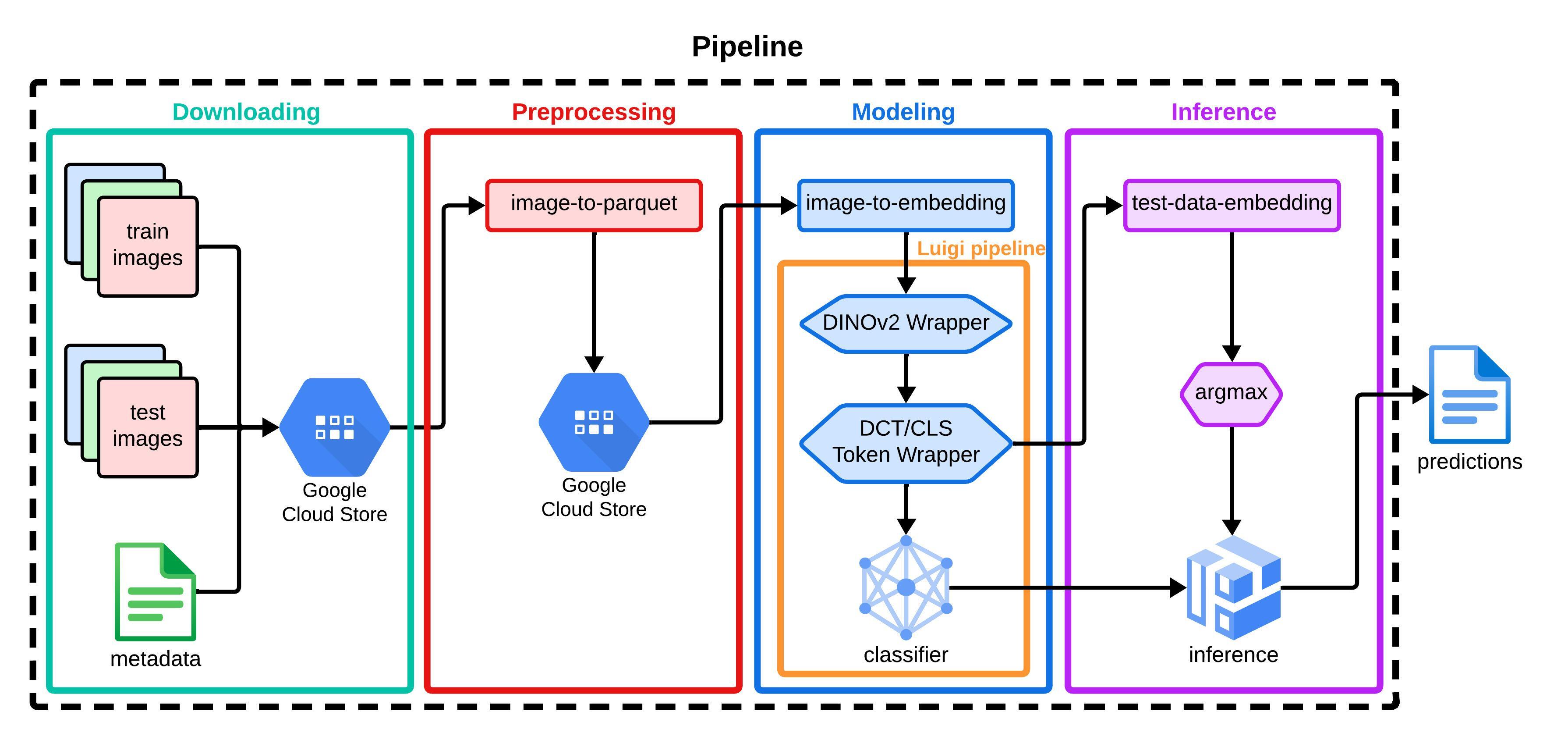}
    \caption{
    End-to-end pipeline.
    The downloading module retrieves the training and test images and the metadata file, storing them in a Google Cloud Storage (GCS) bucket.
    The preprocessing module converts the images to binary data and writes them as parquet files to GCS.
    In the modeling module, the base DINOv2 model extracts embeddings from the training and test data, and a linear classifier is trained on the training embeddings.
    During inference, the trained classifier makes predictions on the test embeddings, formatting the results for leaderboard submission.
    }
  \label{fig:pipeline}
\end{figure}

We trained a classifier to identify snakes using DINOv2 as the primary feature extractor.
The pipeline comprises pre-processing, modeling, and inference stages implemented in separate modules to facilitate the development and testing of the model, as illustrated in Figure \ref{fig:pipeline}.

\subsection{Preprocessing}

Preprocessing includes joining metadata and images into a columnar data store to help aid with data processing in a map-reduce setting.
We use Luigi to orchestrate our pipeline and Apache Spark to process the images and metadata into a Parquet dataset of 200 partitions.
The image binary data is read from the extracted tarballs, and the filename is joined to the corresponding entry in the metadata file.
The species class ID is used as the label in the transfer learning process, which uses the underlying embedding model to predict new tasks and domains.

We select the ViT-B/14 (distilled) base model, which is available via HuggingFace under \texttt{facebook/dinov2-base} pre-trained on LVD-142M.
We pre-compute the DINOv2 embeddings for each image, using the base pre-trained model to transform each image and extract the [CLS] token.
We also keep coefficients from the 2D DCT on patch tokens as an alternative representation.
The 2D DCT maps the embeddings in $\mathbb{R}^{256 \times 768}$ to coefficients in $\mathbb{R}^{256 \times 768}$, of which we keep the top left $8\times8$ coefficients flattened into a vector.
We use the HuggingFace Transformers library with PySpark on a single g2-standard-8 machine on Google Cloud Platform (GCP) using an NVIDIA Tesla L4 GPU.

\subsection{Modeling}

The training module uses the extracted embeddings to train a classifier.
We train a neural network with a linear layer, mapping the embedding space to an output space in the domain of class IDs.
The model is trained with Lightning using a negative log-likelihood (NLL) loss.
We use Adam as the optimizer, and Lightning finds a learning rate automatically.
We use a train/validation split of 80/20 and the hidden test set for the final model evaluation.

The inference module uses the trained model to make predictions on new images.
We build a script to run offline without access to the network.
Due to limitations in the HuggingFace platform, all network access is restricted, including the local loopback network.
As an unfortunate side-effect, PyTorch data loaders must be run in a single process since many concurrency control operations are remote-procedure calls over the network instead of shared memory.
We chain the embedding extraction and model predictions and take the argmax of the output to get the predicted class.
Each observation may have multiple images.
We use the mode of the predictions, using the first image as a tie-breaker for the final prediction.
The first tie-breaker is done for ease of implementation but can also be done randomly or by calculating the mode.

\subsection{Evaluation}

We evaluate our models against the leaderboard hosted on HuggingFace.
The evaluation metrics are the same as the 2023 edition of the competition \cite{joly2023overview}.
The Track 1 metric is the weighted average percentage of the macro-F1 score and weighted accuracy of different types of venomous confusion.
A higher score is better.
The Track 2 metric is a cumulative sum of errors related to identifications, penalizing mistakes such as identifying a venomous snake as a harmless one.
Lower scores are better.
The F1 and Accuracy scores are based simply on model prediction.

\section{Results}

We note the scores for our run in table \ref{tab:results}.
We achieved a score of 39.69 on Track 1, significantly lower than the top score of 85.6 and the baseline score of 67.0.
Track 2 scores are negatively correlated to track 1 scores, with a score of 3790 versus the top score of 687 and the baseline score of 1861.
The model also scored 0 for F1 and Accuracy, meaning most predictions were wrong.

\begin{table}[h!]
\centering
\label{tab:results}
\caption{
    A table capturing the first place, baseline, and DS@GT scores.
}
\begin{tabular}{|c|l|r|r|r|r|l|}
\hline
\textbf{Rank} & \textbf{ID} & \textbf{Track1} & \textbf{Track2} & \textbf{F1} & \textbf{Accuracy} & \textbf{Timestamp} \\ \hline
1 & upupup & 85.63 & 687 & 43.66 & 72.04 & 2024-05-23 20:02:18 \\ \hline
7 & Baseline with Swin-v2 tiny & 67.01 & 1861 & 13.34 & 39.88 & 2024-02-28 08:13:33 \\ \hline
15 & DS@GT-LifeCLEF & 39.69 & 3790 & 0 & 0 & 2024-05-24 21:03:59 \\ \hline
\end{tabular}
\label{tab:submission_rankings}
\end{table}

\section{Discussion}

The model performance is suspiciously bad and could likely be attributed to a bug in implementing label indexing. 
The class IDs are mapped to a contiguous range to avoid training errors in labels but not in the inference code.
This behavior presents as the 0 score for F1 and Accuracy in \ref{tab:results}, where most answers are wrong.
This is consistent with the scores in Track 1 and 2.
In Track 1, the weighted average considers situations where we make harmless predictions.
The score likely reflects the distribution of the data, where there are more venomous than harmless ones.
The Track 2 score is generally high because the model racks up mispredictions.
If the indexing bug were fixed, the performance of our method could likely outperform the baseline.
We believe clear clustering behavior in the DINOv2 embedding space (see Figure \ref{fig:selected-species}) shows potential for future models.

\section{Future Work}

\subsection{Task-specific Modeling}

Our methodology did not consider the custom competition metrics related to venomous classification.
Future methods would include alternative losses that might better guide the optimizer toward solutions that best fit the data distribution and identification constraints.
One of the challenges in a deep learning setting is to find an appropriate loss that can act as a surrogate for our task metric.
In particular, the task metrics are non-differentiable piece-wise functions unsuited for gradient descent.

We would also like to explore alternatives to the negative log-likelihood loss to handle class imbalances without modifying the input data distribution.
The asymmetric loss (ASL) proposed by \cite{ridnik2021asymmetric} handles multi-class imbalances through dynamic down-weighting of easy examples and is amenable to single-class tasks too.

\subsection{Image segmentation}

To learn to perform the image classification task, the machine learning model must learn to extract the signal from each image, which is the most critical factor for differentiating between species. 
Given sufficient data, deep learning models can learn to locate this signal through gradient descent. 
However, the amount of required data depends on the prediction space, variance within data inputs, and model architecture. 
Furthermore, many of the pictures in this dataset only have a small or obstructed portion of the picture in which the snake is present. 
While the model may be able to pick up some signal from the surrounding context of the snake (e.g. plant species frequently associated with the snake species), the most reliable signal will always be the in pixels representing the snake itself. 
To this end, we propose future work that includes methods to isolate this signal from the rest of the image and force the model to learn patterns from this portion.

In particular, we focus on techniques that involve models trained on general image segmentation tasks. 
Some of these models can be used out of the box to perform the task unsupervised, while others involve transfer learning through supervised fine-tuning of pre-trained models. 

\subsubsection{Segment Anything Model (SAM)}

\begin{figure}[htbp]
    \centering
    \begin{minipage}{0.49\textwidth}
        \centering
        \includegraphics[width=\linewidth]{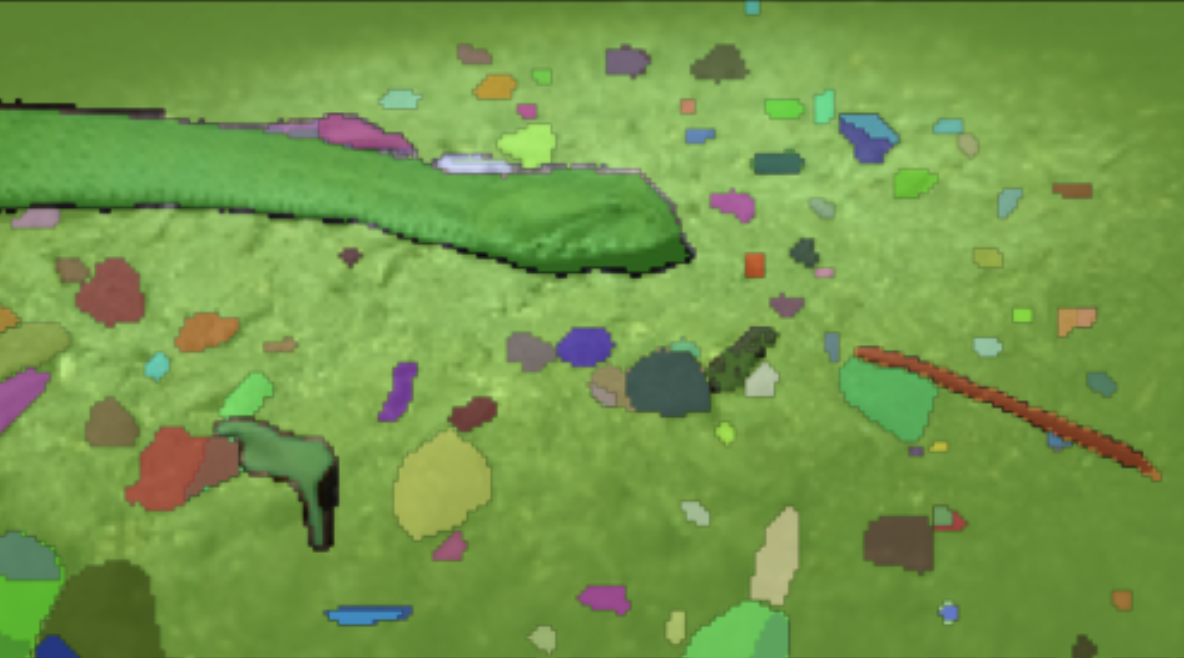}
        \caption{Unsupervised SAM}
        \label{fig:sam-snake}
    \end{minipage}\hfill
    \begin{minipage}{0.49\textwidth}
        \centering
        \includegraphics[width=\linewidth]{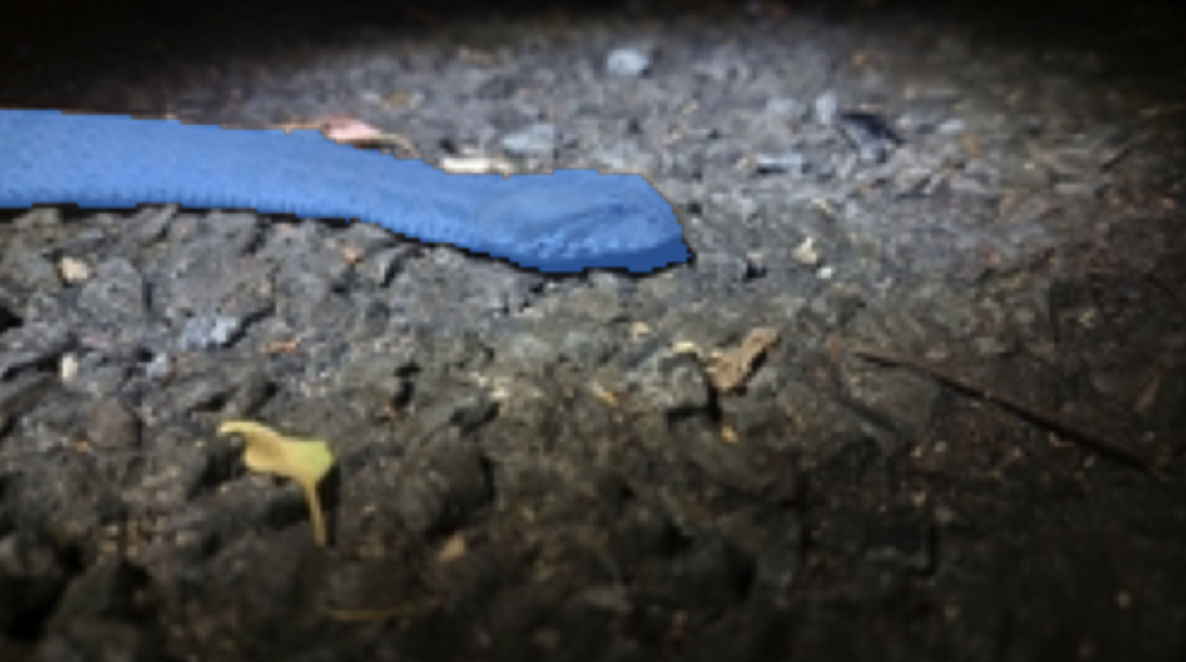}
        \caption{Manually selected SAM segment}
        \label{fig:enter-caption}
    \end{minipage}
\end{figure}

The first model of interest is Meta's Segment Anything Model (SAM) \cite{kirillov2023segment}. 
This model is trained to separate images into distinct segments based on detected objects in the image. 
Although it can be prompted with a point placed on the image, it can also run entirely autonomously and return a set of segments with no human input.
The challenge to using SAM is the inability to determine which returned segments correspond to the snake without further downstream work. 

\subsubsection{OWL-ViT}

OWL-ViT \cite{minderer2022simple} is a pre-trained object detection model that takes candidate label texts and an image as input and returns predicted bounding boxes for the candidate labels' positions in the image. 
Since OWL-ViT can also perform zero-shot inference, it presents a low barrier to use.

In our tests, it labeled snake in only around 44\% of images, most of which were easily discernible to the human eye. 
This level of performance is below our needs for processing incoming images, although it could be helpful in creating a dataset of labeled images from its outputs. 

\subsubsection{Fine-Tuning YOLOv8}

Using supervised learning to train a model on the task of identifying pixels that belong to snakes is perhaps the most promising approach we identified to identify the pixels with the most signal for species classification.
Extra effort is required for this approach, as a labeled dataset of images must be gathered to fine-tune the model for the task. 
Fortunately, open-source datasets are already available for this task, including those found on Roboflow, which contain over 5000 images of snakes labeled for image segmentation. 
Models trained on these datasets have achieved a mean average precision (MAP) of 80.1\% \cite{roboflow-snake-dataset} or higher. 

\begin{figure}[htbp]
    \centering
    \begin{minipage}{0.49\textwidth}
        \centering
        \includegraphics[width=\linewidth]{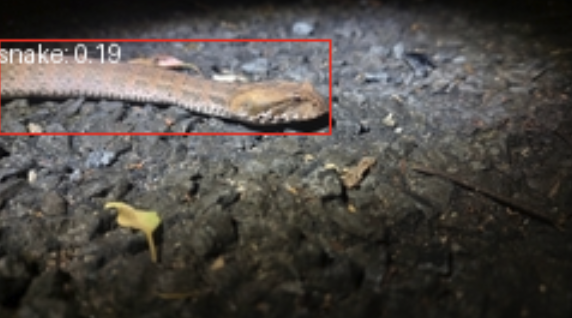}
        \caption{OWL-ViT prediction}
        \label{fig:sam-snake}
    \end{minipage}\hfill
    \begin{minipage}{0.49\textwidth}
        \centering
        \includegraphics[width=\linewidth]{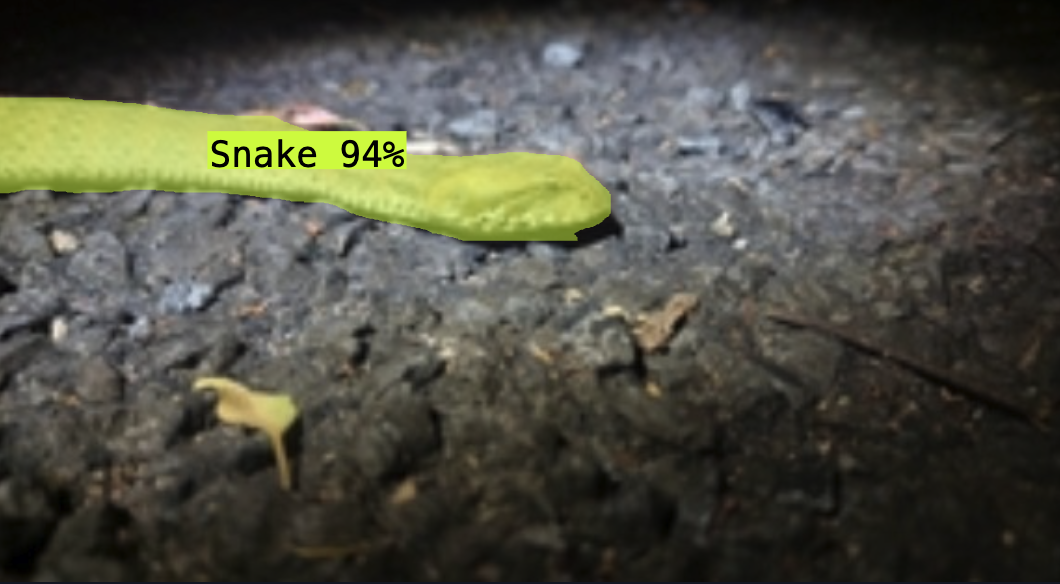}
        \caption{Fine-tuned YOLOv8 prediction}
        \label{fig:enter-caption}
    \end{minipage}
\end{figure}

Using such a model, it is feasible to identify relevant image segments in a high portion of cases.
After successfully identifying the segments of the images containing snakes, these output bounding boxes or masks must be presented to the downstream modeling tasks for embedding and classification. 
In the case of bounding boxes, this can be a straightforward cropping of the image to the box's boundaries. 
This is feasible because DINOv2, our embedding model, accepts images of any aspect ratio. 
In the case of image segmentation masks, there is the possibility of filling the spaces of the image outside the masks with null pixels, preserving only the information identified by our image segmentation model as belonging to a snake. 
While this method allows for the most signal filtering, DINOv2 will need to be evaluated for its performance in handling images with null pixel values, as these are likely outside of its training distribution and could result in unexpected behavior. 

\section{Conclusions}

This paper explored the application of Meta's DINOv2 vision transformer model for snake species identification as part of the SnakeCLEF 2024 competition. 
Our approach involved training a linear classifier on features extracted by DINOv2 from a diverse dataset comprising over 182,000 images. 
Despite a score of 39.69, our work demonstrates the potential of using pre-trained vision models for this species classification task.

Our findings indicate that while the DINOv2 embeddings provide a strong foundation for identifying snake species, there are significant opportunities for improvement. 
These include applying and optimizing more advanced neural network architectures and including image segmentation techniques to isolate relevant features of snakes that could substantially enhance model performance.

Our initial results lay the groundwork for more refined approaches in snake identification using transfer learning. 
We believe significant performance gains can be achieved by addressing current limitations and incorporating advanced segmentation methods, contributing valuable insights to species classification and biodiversity monitoring.
All code for this project is available at \url{https://github.com/dsgt-kaggle-clef/snakeclef-2024}.

\section*{Acknowledgements}

Thank you to the DS@GT CLEF team for providing the hardware for running experiments. Thank you to the organizers of the SnakeCLEF task and LifeCLEF lab for running the competition.

\bibliography{main}


\end{document}